\documentclass[conference,letterpaper]{IEEEtran}

\IEEEoverridecommandlockouts
\ifdefined\pdfminorversion \pdfminorversion=4 \fi

\usepackage[T1]{fontenc}
\usepackage{mathptmx}
\usepackage{amsmath}
\usepackage{amssymb}
\usepackage{graphicx}
\usepackage{booktabs}
\usepackage{multirow}
\usepackage{xcolor}
\usepackage{url}
\usepackage{tabularx}
\usepackage{array}
\usepackage{adjustbox}
\usepackage[hidelinks]{hyperref}

\newcolumntype{C}[1]{>{\centering\arraybackslash}p{#1}}

\definecolor{mConv}{HTML}{8E24AA}   
\definecolor{mSSI}{HTML}{00838F}    
\definecolor{mOmn}{HTML}{558B2F}    
\definecolor{mRiem}{HTML}{795548}   
\definecolor{mNat}{HTML}{455A64}    

\definecolor{dsBCIC2A}{HTML}{0072B2}     
\definecolor{dsPhysioNet}{HTML}{56B4E9}  
\definecolor{dsTUEV}{HTML}{D55E00}       
\definecolor{dsFACED}{HTML}{CC79A7}      
\definecolor{dsMDD}{HTML}{E69F00}        

\title{EEG Foundation Models Need Channel Adaptation, \\
but Fixed Channel Mappings Often Suffice}

\author{\IEEEauthorblockN{Kuntal Kokate\textsuperscript{1}, Dung Truong\textsuperscript{1}, Pierre Guetschel\textsuperscript{2}, Bruno Aristimunha\textsuperscript{1,3}, and Arnaud Delorme\textsuperscript{1,4}}
\IEEEauthorblockA{\textsuperscript{1}Swartz Center for Computational Neuroscience, Institute for Neural Computation,\\
University of California San Diego, La Jolla, CA 92093, USA\\
\textsuperscript{2}Donders Institute for Brain, Cognition and Behaviour, Radboud University, Nijmegen, The Netherlands\\
\textsuperscript{3}Yneuro, Paris, France\quad \textsuperscript{4}Centre de Recherche Cerveau et Cognition, CNRS, Universit\'e Paul Sabatier, Toulouse, France\\
\{kukokate, dutruong, baristimunha, adelorme\}@ucsd.edu\\
pierre.guetschel@donders.ru.nl}}

\IEEEaftertitletext{%
  \noindent\begin{minipage}{\textwidth}
    \footnotesize\raggedright
    \textcopyright{} 2026 IEEE. Personal use of this material is permitted.
    Permission from IEEE must be obtained for all other uses, in any current or
    future media, including reprinting/republishing this material for advertising
    or promotional purposes, creating new collective works, for resale or
    redistribution to servers or lists, or reuse of any copyrighted component of
    this work in other works.
  \end{minipage}\par\vspace{0.6\baselineskip}%
}

\begin{document}

\maketitle
\thispagestyle{empty}
\pagestyle{empty}

\begin{abstract}
Scaling EEG foundation models requires pooling data across heterogeneous montages, so every model needs a channel-adaptation step mapping an arbitrary source montage onto its expected input.
Whether this step must be \emph{learned}, and whether it confounds cross-model comparison, is untested.
We compare five channel-adaptation routes (a learned Conv1d projection, spherical spline interpolation, source-space decomposition, Riemannian re-centering, and each model's native handling) across five pretrained models (5M--157M parameters), five downstream tasks, two training regimes, and up to 15 random seeds.
The answer is architecture-dependent rather than adapter-universal: fixed or native mappings are best in 18 of 25 model--task cells, but regime-specific wins are distributed across Native/Conv1d/SSI/OmnEEG/Riemannian.
Rigid fixed-input BENDR is the main case where a learned bridge is needed; the strongest model, CBraMod, reaches its best results without adding a learned channel adapter.
Because a learned adapter adds a montage-dependent task parameter budget, we recommend reporting frozen-probe benchmarks under a fixed/native route, and treating learned adapters in full fine-tuning as part of the model being evaluated. Source code is available at \url{https://github.com/Kkuntal990/eeg-channel-adaptation}.

\end{abstract}

\section{INTRODUCTION}
A first generation of self-supervised electroencephalography (EEG) foundation models, including BENDR~\cite{kostas2021}, Labram~\cite{labram2024}, Neuro-GPT~\cite{cui2024neurogpt}, EEGPT~\cite{wang2024eegpt}, CBraMod~\cite{wang2025cbramod}, LUNA~\cite{doner2025luna}, and Reve~\cite{reve2025}, has shown transfer across downstream tasks~\cite{banville2021uncovering}.
These models are pretrained on specific electrode montages (e.g., 19-channel 10--20 system or proprietary configurations), and their behavior under realistic montage variation, where downstream datasets routinely use 22, 64, or 21 channels, remains understudied.

\looseness=-1 Every brain foundation model already embeds \emph{some} channel-handling mechanism, from a fixed convolutional front-end (BENDR~\cite{kostas2021}) to learned per-channel projections and embeddings (EEGPT~\cite{wang2024eegpt}, LaBraM~\cite{labram2024}) to coordinate-based positional encoding and cross-attention over electrodes (LUNA~\cite{doner2025luna}). Montage variation is therefore not an unsolved input problem but a choice made at every transfer: preserve the model's pretrained channel computation, or override it with an external, task-trained adapter. 
The override is usually a trainable $1\times1$ convolution borrowed from computer vision, and whether such a learned adapter improves on a model's native handling has never been measured across models. We compare it against three fixed, EEG-specific alternatives:

\begin{itemize}
\item \textbf{\textcolor{mConv}{Learned projection (Conv1d)}}: a trainable $1\times 1$ convolution maps $C_\text{source}$ to $C_\text{target}$ channels, trained jointly with the downstream task~\cite{kostas2021}; the channel-mixing inductive bias is borrowed from computer vision.
\item \textbf{\textcolor{mSSI}{Spatial interpolation (SSI)}}: an EEG-specific, physics-based reconstruction of target channels from source channels using 3D electrode positions and spherical splines~\cite{perrin1989}.
\item \textbf{\textcolor{mOmn}{Source-space decomposition (OmnEEG)}}: signals are projected into a topology-agnostic native representation of 25 spherical harmonic coefficients~\cite{wingeier2001spherical} via the OmnEEG tokenizer~\cite{omneeg2025}, bypassing channel-level mapping.
\item \textbf{\textcolor{mRiem}{Geometric domain adaptation (Riemannian re-centering)}}: per-subject geometric-mean whitening on the SPD manifold aligns source and target distributions~\cite{mellot2024}.
\end{itemize}

\looseness=-1 This preserve-vs-override choice carries three stakes. For \emph{deployment}, it is made at every transfer by habit rather than evidence. For \emph{evaluation}, only the learned projection adds task-trained channel parameters, and its capacity grows with the montage, so a cross-model comparison in which every backbone wears a learned adapter no longer isolates the pretrained representation: it partly ranks adapter capacity. For the \emph{science}, whether native handling suffices tests directly whether pretraining captured transferable spatial structure at all. Yet no systematic comparison of these routes exists across models and tasks: prior work typically evaluates a single adaptation method on a single model, and community benchmarks target cross-task and cross-subject generalization~\cite{aristimunha2025eegchallenge} rather than the adaptation layer itself.
We address this gap with a controlled cross-benchmark spanning five models, five adaptation methods (including \textbf{\textcolor{mNat}{Native}} handling, each model's own built-in channel machinery with no external adapter), five common and largest downstream datasets across four task domains, and two training regimes (probe and supervised fine-tuning, SFT) following the benchmarking methodology of~\cite{guetschel2026openeegbench}.

A natural objection is that channel adaptation is transient: build the model
channel-agnostic from the start (e.g.\ positional encodings from 3D electrode
coordinates), and no external map is needed. The one model built that way
contradicts this. LUNA encodes 3D coordinates and accepts any montage, yet its
\emph{native} handling wins none of its five tasks; a fixed external map is best
on all five, by up to $+13.9$\,pp on depression (OmnEEG 81.9\% vs.\ 68.0\%
native). Channel-\emph{agnostic} (accepting any montage) is thus not
channel-adaptation-\emph{free} (needing no external map); LUNA's own authors
report its coordinate encoding generalizes \emph{worse} to an unseen dense
montage than the coordinate-free CBraMod~\cite{doner2025luna}. Coordinate
encoding fixes geometry, not the distributional shift covariance re-centering
removes, so even a well-designed native handler leaves adaptation on the table.

Our contributions are:
\begin{enumerate}
    \item the first systematic cross-model comparison of channel-adaptation methods, separating a learned projection from four fixed/native routes across five EEG foundation models, five downstream tasks, and two training regimes;
    \item evidence that learned adapters are not a universal default: a fixed mapping or native handling is best in 18 of 25 model--task cells, while method wins split across all five routes when probe and SFT are counted separately;
    \item a rigid/flexible account of when adaptation is nonetheless required, with an architecture-dependent optimum and a probe-SFT asymmetry consistent with small-data feature distortion during fine-tuning under a shared learning rate; and
    \item a benchmarking recommendation: frozen-probe comparisons should use a fixed/native route to avoid adapter-capacity confounds, whereas full-SFT comparisons should report the adapter as part of the trainable system.
\end{enumerate}

\section{RELATED WORK}

\subsection{EEG Foundation Models}

Recent EEG foundation models pretrain on large unlabeled corpora with self-supervised objectives: masked reconstruction (LaBraM~\cite{labram2024}, CBraMod~\cite{wang2025cbramod}, EEGPT~\cite{wang2024eegpt}), contrastive learning (BENDR~\cite{kostas2021}), and autoregressive modeling (Neuro-GPT~\cite{cui2024neurogpt}, NeuroLM~\cite{neurolm2025}). They differ mainly in how they ingest channels. We evaluate five that span that axis: BENDR (convolutional encoder, fixed 20 channels), Neuro-GPT (convolution plus GPT-2, fixed 22), EEGPT (patch transformer, learned per-name channel vocabulary), CBraMod (criss-cross transformer, coordinate-free), and LUNA (cross-attention with 3D-coordinate queries, topology-agnostic). They are chosen to span rigid-to-flexible channel handling with weights loadable in one framework rather than to maximize scale; Sec.~\ref{sec:native} reduces their handling to three primitives.

\subsection{Channel Adaptation for EEG}

Prior work populates the four families introduced above. \textbf{\textcolor{mConv}{Learned projection}} trains a channel-mixing map jointly with the task, as in BENDR~\cite{kostas2021} and EEGPT~\cite{wang2024eegpt}. \textbf{\textcolor{mSSI}{Spatial interpolation}} reconstructs target channels from reference electrodes via spherical splines~\cite{perrin1989} or forward head models~\cite{mellot2024}. \textbf{\textcolor{mOmn}{Source-space decomposition}} projects EEG onto a topology-agnostic harmonic basis~\cite{wingeier2001spherical} atop Yao's spherical standardization~\cite{yao2001rest}, implemented by the OmnEEG tokenizer~\cite{omneeg2025}; its effectiveness relative to simpler methods is unestablished. \textbf{\textcolor{mRiem}{Geometric domain adaptation}} aligns per-subject distributions on the SPD manifold via Riemannian re-centering~\cite{mellot2024, barachant2012}.

Some models instead fold channel handling into pretraining itself, through cross-attention over electrodes~\cite{doner2025luna} or coordinate-based positional encoding~\cite{reve2025}. Rather than setting these architectural mechanisms aside, we treat each model's built-in handling as its \textbf{\textcolor{mNat}{Native}} route, making the ``preserve the pretrained handling vs.\ override it'' question testable on flexible and rigid backbones alike. More broadly, brain-computer interface (BCI) transfer learning~\cite{sartzetaki2023} typically assumes fixed electrode montages, leaving the channel adaptation component understudied.

\section{METHODS}

\subsection{Channel adaptation as an attribution problem}

All four families considered here apply a linear map along the channel axis to transform the source signal $\mathbf{X}_s \in \mathbb{R}^{C_s \times T}$ into the representation $\mathbf{X}_t \in \mathbb{R}^{C_t \times T}$ expected by the foundation model:
\begin{equation}
\mathbf{X}_t = \mathbf{M}\,\mathbf{X}_s,
\label{eq:unified}
\end{equation}
where $\mathbf{M} \in \mathbb{R}^{C_t \times C_s}$ is the \emph{adaptation matrix}. The important control is not only how $\mathbf{M}$ is constructed, but whether it is learned from downstream labels. In a frozen probe, a learned channel map is extra task capacity; in SFT, it is one more trainable component in a fully trainable system. Table~\ref{tab:adapter_accounting} makes this accounting explicit.

\subsubsection{\textbf{\textcolor{mConv}{Learned projection (Conv1d)}}}
$\mathbf{M}$ is a trainable $1\times 1$ convolution (with optional bias), learned end-to-end with the downstream task. Its inductive bias is the simplest of the four: no electrode positions are needed, and the only prior is that channels mix linearly.

\subsubsection{\textbf{\textcolor{mSSI}{Spatial interpolation (SSI)}}}
$\mathbf{M}$ is a fixed interpolation matrix computed analytically from 3D electrode coordinates using Legendre-polynomial spherical splines~\cite{perrin1989}, via MNE-Python~\cite{gramfort2013}. We target the model-compatible canonical montage used by each wrapper. BENDR expects 20 channels. Its Conv1d route is SSI followed by a learned $19\!\rightarrow\!20$ $1\times1$ bridge, its OmnEEG route uses a learned $25\!\rightarrow\!20$ bridge, and its SSI and Riemannian routes feed the 19 channels into a newly initialized first encoder layer. Every BENDR route is therefore hybrid (Table~\ref{tab:adapter_accounting}).

\subsubsection{\textbf{\textcolor{mOmn}{Source-space decomposition (OmnEEG)}}}
$\mathbf{M} \in \mathbb{R}^{25\times C_s}$ is a fixed basis matrix whose rows evaluate the real spherical harmonics $Y_{\ell m}(\theta_i,\phi_i)$ at the source electrode positions, for orders $\ell\leq 4$~\cite{wingeier2001spherical, omneeg2025}. The output is a topology-agnostic 25-coefficient representation that does not depend on $C_s$.

\subsubsection{\textbf{\textcolor{mRiem}{Geometric domain adaptation (Riemannian re-centering)}}}
Applied after SSI, $\mathbf{M} = \bar{\mathbf{C}}_j^{-1/2}$ is a per-subject whitening on the SPD manifold, where $\bar{\mathbf{C}}_j$ is the geometric mean of subject $j$'s unlabeled covariance matrices (with Ledoit-Wolf shrinkage)~\cite{mellot2024, barachant2012}. The full pipeline therefore applies $\bar{\mathbf{C}}_j^{-1/2}\,\mathbf{G}$ to $\mathbf{X}_s$, aligning per-subject distributions without using task labels.

\begin{table}[t]
\caption{Adapter accounting used for attribution. $C_m$ denotes the channel representation consumed by the model wrapper.}
\label{tab:adapter_accounting}
\centering
\scriptsize
\setlength{\tabcolsep}{2.5pt}
\renewcommand{\arraystretch}{0.92}
\begin{tabular}{p{0.15\columnwidth}p{0.54\columnwidth}p{0.22\columnwidth}}
\toprule
\textbf{Route} & \textbf{Fitted before model} & \textbf{Learned channel params.} \\
\midrule
\textcolor{mNat}{Native} & No external map; pretrained channel handler frozen in probe, trainable in SFT. & 0 external$^\dagger$ \\
\textcolor{mConv}{Conv1d} & Direct $C_s\!\rightarrow\!C_m$ map fit from labels. & $C_sC_m$ (+ bias) \\
\textcolor{mSSI}{SSI} & Fixed spherical-spline map from electrode coordinates. & 0$^\dagger$ \\
\textcolor{mRiem}{Riem.} & SSI plus unlabeled per-subject covariance re-centering. & 0$^\dagger$ \\
\textcolor{mOmn}{OmnEEG} & Fixed $C_s\!\rightarrow\!25$ spherical-harmonic basis. & 0$^\dagger$ \\
\bottomrule
\multicolumn{3}{p{0.94\columnwidth}}{\footnotesize $^\dagger$Exceptions: all BENDR routes (see text); learned $1\times1$ bridges after Neuro-GPT's external routes and after OmnEEG for EEGPT and CBraMod; EEGPT's Native route is a learned $C_s\!\rightarrow\!19$ projection.}
\end{tabular}
\end{table}

\subsection{Foundation Models}

Table~\ref{tab:models} summarizes the five models, spanning 5M--157M parameters and rigid-to-flexible channel handling. Pretrained weights are loaded from HuggingFace Hub via braindecode~\cite{braindecode} (BENDR, EEGPT, LUNA, CBraMod) and the authors' release (Neuro-GPT).
\begin{table}[t]
\caption{EEG Foundation Models Evaluated}
\label{tab:models}
\centering
\small
\begin{adjustbox}{max width=\columnwidth}
\begin{tabular}{lcccc}
\toprule
\textbf{Model} & \textbf{Params} & \textbf{Architecture} & \textbf{Ch.} & \textbf{sfreq} \\
\midrule
BENDR~\cite{kostas2021} & 157M$^\dagger$ & Conv Encoder & 20 & 256 Hz \\
Neuro-GPT~\cite{cui2024neurogpt} & 79.5M & Conv + GPT-2 & 22 & 256 Hz \\
EEGPT~\cite{wang2024eegpt} & 25.3M & Patch Transformer & Any & 256 Hz \\
LUNA~\cite{doner2025luna} & 7.1M & Cross-Attention & Any & 256 Hz \\
CBraMod~\cite{wang2025cbramod} & 5.0M & Criss-Cross Trans. & Any & 200 Hz \\
\bottomrule
\multicolumn{5}{l}{\footnotesize $^\dagger$Total; encoder and contextualizer both used.}
\end{tabular}
\end{adjustbox}
\end{table}

\subsection{Native channel handling}
\label{sec:native}
Backbone labels (``Conv Encoder'', ``Patch Transformer'', ``Cross-Attention'') hide what actually processes channels. Two questions separate the mechanisms: how a model \emph{identifies} a channel, and how it \emph{combines} channels. Along these two axes the five backbones reduce to three families. \textbf{F1, a learned linear filter:} BENDR's first 1D convolution and Neuro-GPT's $(C,1)$ spatial convolution are the same operation, a fixed-size linear map $\mathbb{R}^{C}\!\to\!\mathbb{R}^{d}$ that bakes the channel order into its weights and cannot accept a new montage without an external map. \textbf{F2, learned-query pooling:} EEGPT's summary tokens and LUNA's learned queries are the same bottleneck, a small set of learned queries that pools any number of channels; they differ only in how they identify a channel (EEGPT by a discrete learned name embedding, LUNA by a continuous 3D coordinate). \textbf{F3, conditional encoding with self-attention:} CBraMod attends over all channels and adds a positional encoding generated on the fly by a convolution (ACPE), avoiding coordinates because a channel's identity depends on its reference (as in TUEV's average-vs-linked-ear montages). The family predicts which external adapter fits: one that is the same kind of operation as the native handler composes, while a foreign one competes. A learned $1\times1$ Conv1d is linear, so it composes with F1 (landing in BENDR's pretrained feature space) and is subsumed by EEGPT's own internal $1\times1$ filter; placed before an attention-based handler it is a foreign operation and conflicts, the source of the negative-transfer cases in Sec.~\ref{sec:results}.

\subsection{Training Regimes}

We evaluate two training regimes:
\begin{itemize}
\item \textbf{Probe (frozen encoder)}: Only the classification head and any task-trained channel adapter are trained while pretrained parameters remain frozen. This is the cleanest representation-quality comparison because fixed/native routes keep the trainable budget fixed.
\item \textbf{SFT (supervised fine-tuning)}: All model parameters are unfrozen and trained end-to-end but with a lower learning rate. In this regime the adapter is part of the trainable system, so SFT measures deployable recipe performance rather than isolated representation quality.
\end{itemize}

\subsection{Datasets}

\begin{table}[t]
\caption{Downstream Evaluation Datasets. Channel counts match the harness dataset configurations.}
\label{tab:datasets}
\centering
\small
\begin{adjustbox}{max width=\columnwidth}
\begin{tabular}{lccccl}
\toprule
\textbf{Dataset} & \textbf{Classes} & \textbf{Ch.} & \textbf{Samples} & \textbf{Chance} & \textbf{Spatial char.} \\
\midrule
BCIC2A    & 4 & 22 & 5,184  & 25.0\% & Focal \\
PhysioNet & 4 & 64 & 9,838  & 25.0\% & Focal \\
TUEV      & 6 & 21 & 1,047  & 16.7\% & Mixed \\
\midrule
FACED     & 9 & 26 & 19,217 & 11.1\% & Distributed \\
MDD       & 2 & 19 & 3,857  & 50.0\% & Global \\
\bottomrule
\end{tabular}
\end{adjustbox}
\end{table}

\begin{figure*}[t]
\centering
\includegraphics[width=\textwidth]{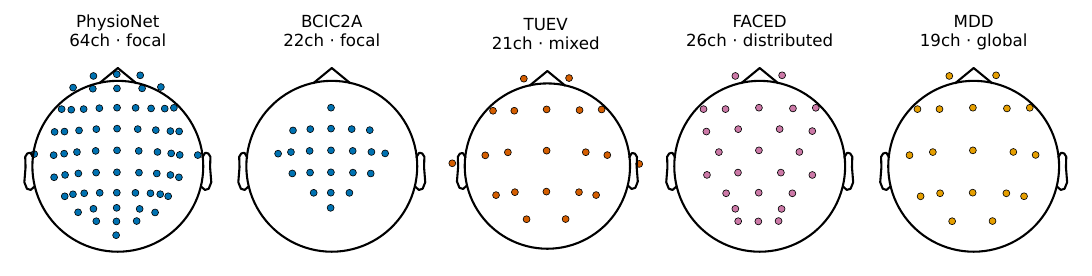}
\caption{\textbf{The five downstream montages span the spatial-frequency axis.}
Electrode layout per dataset (nose up), ordered by task spatial character
(colour; Table~\ref{tab:datasets}): focal sensorimotor motor imagery (dense
PhysioNet, partial-head BCIC2A), mixed clinical events (TUEV), distributed
emotion (FACED), and the global depression biomarker (MDD).}
\label{fig:montages}
\end{figure*}

The five datasets (Table~\ref{tab:datasets}) span four task domains (motor imagery, clinical event detection, emotion recognition, and resting-state depression screening) with diverse channel counts and subject populations (Fig.~\ref{fig:montages}).

The first, \emph{BCI Competition IV 2a (BCIC2A)}~\cite{brunner2008}, has 9 subjects with four motor-imagery classes on a 22-channel central sensorimotor strip (no temporal/occipital coverage), split 1--3/4--6/7--9. The second, \emph{PhysioNet Motor Imagery}~\cite{schalk2004}, has 109 subjects on a dense 64-channel whole-head 10--10 grid, split 1--70/71--89/90--109. The third, \emph{TUH EEG Events (TUEV) v2.0.1}~\cite{obeid2016}, is a clinical corpus of six event types (epileptiform, slowing, artifact, background) on a sparse 21-channel 10--20 montage whose channel meaning is reference-dependent (average vs.\ linked-ear; 40+ corpus configurations), using the official 812/235 split. The fourth, \emph{FACED}~\cite{chen2023faced}, has 123 subjects viewing emotional videos on a 26-channel whole-head montage (common-average reference), nine affective classes in 10-second windows, split 0--79/80--99/100--122. Finally, \emph{MDD Mumtaz}~\cite{mumtaz2017} is a 2-class depression screening task from 64 subjects on a 19-channel 10--20 montage, linked-ear referenced, eyes-closed resting (5-second windows), with predefined splits.

\subsection{Experimental Setup}

All experiments use AdamW optimizer with cosine annealing and warmup (5 epochs). Training runs for up to 50 epochs with early stopping (patience=10, monitoring validation loss). Normalization is model-specific to match pretraining: BENDR and LUNA use min-max scaling to $[-1, 1]$; EEGPT uses channel-wise z-score; CBraMod uses $\mu$V/100 scaling. Each model's data is resampled to its pretrained frequency (Table~\ref{tab:models}). The selection of preserving the normalization from the original authors for models was based on experience, and not replicability components reported by \cite{guetschel2026openeegbench}.

For probe mode, only the classification head (and Conv1d projection, if applicable) is trained. For SFT, all parameters are unfrozen. We report macro-averaged test accuracy (equivalent to balanced accuracy; mean $\pm$ std over up to 15 random seeds) on held-out test splits. A single learning rate per regime was used for all models; per-model tuning may yield further gains. Pairwise significance was assessed via Wilcoxon signed-rank tests on matched seed-level accuracies with Benjamini-Hochberg (BH) correction for 50 best-vs-runner-up tests (FDR $<$ 0.05), with effect sizes via Cohen's $d$ and overall method differences via a Friedman test. Because subject splits are fixed, these tests quantify optimizer-seed stability under the benchmark protocol rather than population uncertainty over new subject partitions.

\section{RESULTS}
\label{sec:results}

\subsection{Fixed/native routes match or beat learned Conv1d in 18 of 25 cells}

Applying a pretrained foundation model to a new montage forces a channel-adaptation choice, but no single adapter wins overall. The choice is per-backbone: whether a backbone needs a task-trained channel projection, or whether a fixed/native route already lands the signal in the representation space that pretraining made useful.

\begin{figure}[!t]
\centering
\includegraphics[width=\columnwidth]{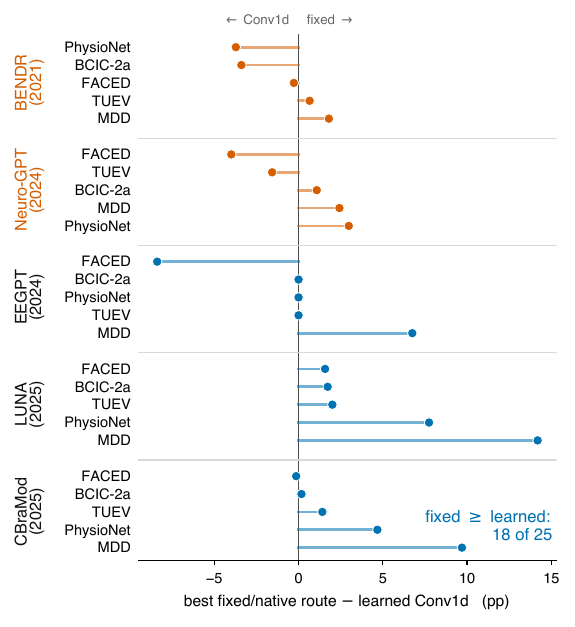}
\caption{\textbf{Fixed/native routes usually match or beat learned Conv1d.} Each row is one backbone $\times$ dataset cell; the lollipop is the best fixed/native route (Native/SSI/OmnEEG/Riemannian) minus learned Conv1d, maxed over probe and SFT. Rightward or zero means fixed/native matches or wins, 18 of 25 cells; leftward bars cluster in the fixed-input backbones (BENDR, Neuro-GPT) and near-chance FACED. For BENDR every route contains learned or re-initialized channel layers (Table~\ref{tab:adapter_accounting}); Fig.~\ref{fig:winner_heatmap} gives regime-specific winners.}
\label{fig:main_results}
\end{figure}

\begin{figure*}[!t]
\centering
\includegraphics[width=\textwidth]{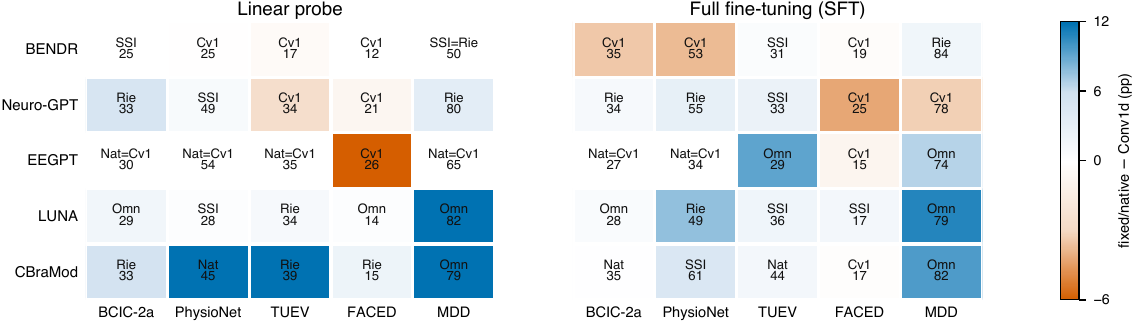}
\caption{\textbf{The winning adapter changes with the training regime.} Winning route and balanced accuracy per backbone $\times$ dataset, split by frozen probe and full SFT. Colour matches Fig.~\ref{fig:main_results}: blue where the best fixed/native route beats learned Conv1d, orange where Conv1d wins; ``='' marks ties. No adapter family dominates the 50 cells, yet Conv1d wins concentrate in fixed-input backbones and FACED while CBraMod keeps native/fixed preprocessing.}
\label{fig:winner_heatmap}
\end{figure*}

Fig.~\ref{fig:main_results} encodes this as a signed gap for every model vs dataset cell: the best fixed/native route (Native, \textbf{\textcolor{mSSI}{Spatial interpolation}}, \textbf{\textcolor{mOmn}{Source-space decomposition}}, or \textbf{\textcolor{mRiem}{Geometric domain adaptation}}) minus the \textbf{\textcolor{mConv}{Learned projection}} route, the traditional computer vision-inspired option, each allowed to use its better training regime. Rightward bars, therefore, mean that no direct learned Conv1d projection was needed. This is the majority case: fixed/native routes match or beat learnable projection in 18 of 25 cells, while the seven Conv1d wins cluster in the fixed-input backbones and in FACED, where all methods remain close to failure (Fig.~\ref{fig:winner_heatmap}).

This majority result should not be read as a universal adapter ranking. When probe and SFT are counted separately (Fig.~\ref{fig:winner_heatmap}), Native/Conv1d/SSI/OmnEEG/Riemannian win 3/13/8/9/10 of the 50 cells, with 7 ties. The differences are often small: the Friedman test is significant under SFT ($p{=}0.012$) but not under the probe ($p{=}0.21$), only 16 of 50 best-vs-runner-up comparisons survive BH correction (15 with $|d|\geq0.8$), and 43.5\% of method--condition pairs fall within 1pp of the best. The finding is therefore about attribution: adapter choice matters where it changes the computation the backbone sees, and is often interchangeable once that computation is preserved.

\subsection{Fixed-input backbones are the recurring Conv1d cases}
\looseness=-1 BENDR and Neuro-GPT expect fixed layouts (20 and 22 channels) and cannot read a new montage without external adaptation; the question is which adapter each prefers. For BENDR it is the learned Conv1d bridge: its frozen-encoder probe sits at chance for every method (${\approx}25\%$ on the four-class tasks), so BENDR is usable only under fine-tuning, where Conv1d leads on the three non-clinical tasks (34.6\% BCIC2A, 53.1\% PhysioNet, 19.4\% FACED), beating spherical interpolation on BCIC2A ($d{=}0.84$). Its two clinical tasks instead prefer a geometry-preserving map (Riemannian 84.0\% MDD; SSI 30.5\% TUEV). Neuro-GPT, whose spatial convolution learned physically meaningful patterns from 10--20 data, prefers physics- and geometry-preserving adaptation on the spatially structured tasks (Riemannian reaches 54.5\% on PhysioNet vs.\ 51.5\% Conv1d, and 34.3\% on BCIC2A), while a learned Conv1d recovers the lead on the two least spatial tasks (34.4\% TUEV probe, $d{=}1.39$; 24.6\% FACED SFT, $d{=}3.11$). In both models the winning adapter is the one closest to the pretrained input layer; we give the mechanism in Sec.~\ref{sec:whyarch}.

\subsection{Flexible backbones mostly use their native channel machinery}
EEGPT, LUNA, and CBraMod accept arbitrary channel counts natively, but they do not behave identically. EEGPT reaches its top accuracy natively on three of five tasks (29.8\% BCIC2A, 53.6\% PhysioNet, 35.2\% TUEV), where probe Conv1d is seed-identical to native; on MDD, native and external Conv1d coincide (64.6\% probe) and source-space decomposition gives its best result (74.2\% SFT). CBraMod (5M) reaches its best with a fixed/native route on four of five tasks (native on BCIC2A and TUEV, SSI on PhysioNet, and source-space decomposition on MDD, 81.7\%), using learned Conv1d only on FACED. Its native probe reaches 44.7\% on PhysioNet against 32.7\% for Conv1d probe ($d{=}5.09$), so its pretrained channel/position handler is not improved by a learned linear remapping in this setting.

LUNA is the informative exception among flexible models: native wins none of its five tasks and external fixed maps win every one: OmnEEG 81.9\% on MDD probe ($d{=}7.37$ vs.\ Riemannian), Riemannian 48.7\% on PhysioNet SFT (+9.3pp over native SFT). Its cross-attention bottleneck re-attends to whatever input it is given rather than committing to a montage prior, so an external map complements rather than competes with it (Sec.~\ref{sec:whyarch}).

\subsection{Fine-tuning can erase a strong probe representation}

\begin{figure}[t]
\centering
\includegraphics[width=\columnwidth]{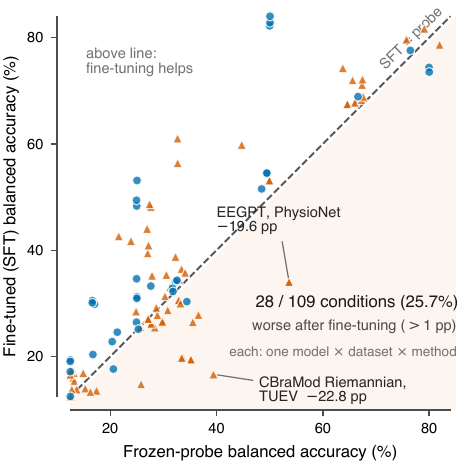}
\caption{\textbf{Fine-tuning is not a monotone improvement over probing.} Each marker is one condition ($N{=}109$); axes give probe ($x$) vs.\ full-SFT ($y$) accuracy. \textcolor[HTML]{0072B2}{$\bullet$}~rigid backbones (BENDR, Neuro-GPT); \textcolor[HTML]{D55E00}{$\blacktriangle$}~flexible (EEGPT, LUNA, CBraMod). Below the shaded $1$\,pp band: $28/109$ ($25.7\%$), chiefly flexible backbones. The largest drops (EEGPT native PhysioNet $-19.6$\,pp, CBraMod\,+\,Riemannian TUEV $-22.8$\,pp) show SFT distorting a useful representation.}
\label{fig:probe_sft}
\end{figure}

Adapter choice is only one source of variation; the training regime can reverse the conclusion within the same model--adapter pair. Full fine-tuning is not uniformly better than a frozen probe: 28 of 109 conditions (25.7\%) lose more than 1pp under SFT (Fig.~\ref{fig:probe_sft}). The regressions are model-specific rather than a rigid/flexible split. Rigid backbones, which must be taught the new mapping, always gain (BENDR mean $+17.0$pp, 0 of 15 conditions regress). Among flexible backbones the two most alike behave oppositely: EEGPT regresses in 19 of 24 conditions (mean $-5.1$pp), with its worst drops on the \emph{native} path (PhysioNet $-19.6$pp, TUEV $-15.7$pp), whereas CBraMod \emph{improves} in 19 of 25 (mean $+5.9$pp) and prefers native SFT. LUNA gains from external adaptation under SFT (mean $+3.6$pp), and the one severe flexible collapse is CBraMod\,+\,Riemannian on TUEV ($-22.8$pp). Because two equally flexible models diverge, the effect tracks the backbone, not montage flexibility; we attribute it to feature distortion under the shared learning rate.

\subsection{Source-space decomposition is a global-biomarker adapter}
One method's usefulness is set by the task, not the model. Source-space decomposition gives the best MDD result for both topology-agnostic backbones (LUNA 81.9\% probe, +16pp over native; CBraMod 81.7\% SFT) and lifts even BENDR above chance on clinical data (28.7\% TUEV and 77.1\% MDD under SFT), while staying at chance on motor imagery (${\approx}25\%$) and near the empirical FACED collapse floor (${\approx}12.5\%$, close to the nominal 11.1\% chance). As a spatial low-pass it suits depression's distributed biomarker and fails where the signal is focal (Sec.~\ref{sec:omneeg_disc}).

\subsection{Architecture, not scale or mismatch size, sets the ceiling}

The regime-specific grid in Fig.~\ref{fig:winner_heatmap} also shows that model scale does not predict downstream performance. The 5M-parameter CBraMod achieves the best results on three of five datasets: BCIC2A (35.4\%), PhysioNet (61.0\%), and TUEV (44.0\%). EEGPT leads on FACED (25.8\% via Conv1d probe), a task where all models struggle, and the 157M BENDR attains the study's highest single accuracy on MDD (84.0\% via Riemannian re-centering). Our five models confound architecture with pretraining data (CBraMod uses TUEG, the largest corpus), training objective, and sampling rate. 

CBraMod's advantage is consistent with independent findings~\cite{eegfmworth2026} that compact architectures can match larger ones, though its inductive bias, TUEG data, and objective cannot be isolated here. Nor does the size of the montage mismatch predict how much adaptation helps: across all model-dataset pairs, the accuracy gained by the best adaptation method is essentially uncorrelated with the number of mismatched channels (Spearman $\rho=-0.07$), so a 44-channel gap (PhysioNet into BENDR's 20-channel input) is no more amenable to adaptation than a 1-channel gap (TUEV into BENDR). Architecture, not mismatch size, sets the ceiling.

\section{DISCUSSION}

\subsection{One principle: preserve the pretrained channel computation, don't overwrite it}
\label{sec:whyarch}

One rule organizes the pattern: a backbone's best adapter is usually the one whose output lands where its pretrained channel-handling component already expects signal, so adaptation should \emph{preserve} that computation rather than \emph{overwrite} it. What ``preserve'' means is set by the architectural component that ingests channels, which is why the optimum tracks the model more than parameter count or the number of missing electrodes.

\textbf{BENDR's first Conv1d layer} mixes all channels into 512 feature maps tuned to the joint statistics of its fixed 20-channel input. A learned Conv1d bridge optimizes directly into that feature space (why it beats physics-based methods on the non-clinical tasks), whereas our SSI and Riemannian routes replace that first layer with a newly initialized one. Its frozen probe is at chance for every method, so the features are not linearly usable off the shelf; and on depression Riemannian re-centering overtakes it (84.0\% vs.\ 82.2\%), subject-level covariance alignment helping a broad resting-state biomarker but not focal motor features.

\textbf{Neuro-GPT's spatial convolution} applies a (C=22, 1) kernel trained on 10--20 data; SSI/Riemannian preserve that geometry and win on the spatially structured tasks, while on TUEV and FACED, where the signal is less tied to sensorimotor topography, Conv1d recovers the lead.

\looseness=-1 \textbf{The flexible backbones} carry their channel prior in different components: EEGPT's per-electrode embedding codex, which subsumes an external Conv1d in several cells; CBraMod's content-conditioned ACPE; and LUNA's learned-query bottleneck. For CBraMod, native handling is already a pretrained montage-to-representation map, so an external adapter often adds a competing representation rather than useful capacity. LUNA is the exception that clarifies the mechanism: its bottleneck re-attends to whatever input it is given, so an external map complements the model instead of colliding with a fixed native prior.

The rigid--flexible split is therefore about channel identity, not scale. The F1 convolutions bind weights to a channel order and dimensionality, so a changed montage is an out-of-distribution object at the first layer; the F2/F3 backbones make position explicit (CBraMod's conditional ACPE~\cite{chu2021conditional, wang2025cbramod}, LUNA's coordinate encoding~\cite{doner2025luna}) and escape that binding (Sec.~\ref{sec:native}). Our evidence is indirect (accuracy and fine-tuning behavior, not a direct spatial-information probe), but it predicts the pattern: native handling works when the pretrained channel computation already matches the montage, and external maps help when it does not.

\subsection{The probe--SFT asymmetry is feature distortion, not adapter conflict}
\label{sec:probe_sft_disc}

\looseness=-1 Fine-tuning falls below a frozen probe in 28 of 109 conditions (25.7\%), and the regressions are model-specific: EEGPT loses ground in 19 of 24 (mean $-5.1$pp) while the equally flexible CBraMod improves in 19 of 25 (mean $+5.9$pp). The natural explanation, that an external adapter fights the internal channel handler, is incomplete, because EEGPT's largest drops are on its \emph{native} path, where no external adapter exists (PhysioNet $53.6\rightarrow34.0$, TUEV $35.2\rightarrow19.5$); the pretrained representation itself is degraded. Kumar et al.~\cite{kumar2022} identify the mechanism: full fine-tuning updates a randomly-initialised head and the backbone jointly, so early head gradients distort well-pretrained features, worst when the representation is already near-optimal and the downstream set is small: the regime here. Lee et al.~\cite{lee2023surgical} reach the same conclusion from the other side, showing that tuning only the layers matched to the shift beats full fine-tuning while small-data fine-tuning forgets pretraining; we evaluate only probe and full SFT, so we observe the distortion but apply neither remedy. Whether distortion dominates depends on the backbone: EEGPT's dual objective is the most fine-tuning-fragile in our set, whereas CBraMod's masked-patch objective tolerates and often prefers fine-tuning. LUNA is again the exception (its query bottleneck stays stable and gains under SFT), and the CBraMod Riemannian collapse on TUEV ($39.4\rightarrow16.7$) is the clearest adapter-conflict case. Both mechanisms operate; the native-path EEGPT failures show that adapter conflict alone cannot explain the asymmetry.

\subsection{\textbf{\textcolor{mOmn}{Source-space decomposition}} is a global-biomarker tool, not a general adapter}
\label{sec:omneeg_disc}

Source-space decomposition ($\ell\leq 4$, 25 coefficients) acts as a spatial low-pass filter: it preserves global patterns such as hemispheric asymmetries and anterior--posterior gradients, but it cannot represent focal activity carried by higher spatial orders. This predicts where it helps. It succeeds on MDD, where reported depression biomarkers are broad spectral/connectivity patterns~\cite{mumtaz2017}, and fails on motor imagery, where discriminative $\mu/\beta$ desynchronisation is focal. Its partial success with rigid models on MDD and TUEV suggests that coarse clinical structure can survive even mismatched pretrained filters when the task tolerates low spatial resolution. The account is mechanistic but not directly measured; source-space is therefore a domain-specific clinical tool, not a general adapter. REVE's 4D positional encoding~\cite{reve2025} marks the complementary path: build montage flexibility into pretraining rather than retrofitting it afterwards.

\subsection{Practical guidelines and the benchmarking stake}

\textbf{Decision framework:} Default to the least adaptive route that preserves the pretrained channel computation, and add a learned adapter only when the backbone cannot otherwise accept the montage. (1)~For flexible models (EEGPT, LUNA, CBraMod), use native handling or a fixed mapping; a learned Conv1d adds complexity and, under full SFT, may amplify negative transfer. LUNA is the exception, benefiting from SSI/Riemannian because its bottleneck can exploit an external map. (2)~For Neuro-GPT, use SSI/Riemannian when the task is spatially structured, because these routes preserve the 10--20 geometry its spatial convolution expects. (3)~BENDR is the sole case that routinely requires a learned projection, because its first convolution cannot otherwise accept a mismatched montage; even then, Riemannian is preferable on clinical MDD.

\textbf{Benchmarking implication:} A learned adapter adds a montage-dependent parameter budget, so comparing frozen encoders with their own learned adapters conflates representation quality with adapter capacity. For representation benchmarking, use one fixed/native route and a frozen encoder; accuracy differences then more cleanly reflect the pretrained representation under a shared input transform. For full SFT, the claim is weaker but still useful: a fixed adapter removes one capacity confound, while the trainable backbone remains part of the evaluated recipe. Under the stricter frozen-probe control, the best result on three of five datasets uses no learned channel adapter, which we propose as a desideratum for montage-robust EEG foundation models.

\textbf{Where adaptation lives.} The budget asymmetry reflects \emph{where} adaptation sits: the fixed routes (SSI, Riemannian, source-space) are external, in the preprocessing-alignment lineage of Euclidean Alignment~\cite{hewu2019euclidean}, whereas a learned Conv1d is an \emph{in-model} adapter whose size scales with channel count~\cite{houlsby2019adapter, hu2021lora}. A native codex or ACPE is in-model too but fixed at pretraining, so only an added learned adapter grows the budget with the montage: the rigid--flexible contrast is whether channel invariance is \emph{built into} the architecture or \emph{learned onto} a fixed one~\cite{benton2020learning}.

\textbf{Compute trade-offs:} SSI is offline (zero training overhead), Conv1d adds a small learnable layer, Riemannian needs per-subject covariance estimation ($O(C^2 T)$), and source-space needs electrode positions and a harmonic projection; when Conv1d and SSI tie, prefer SSI for simplicity.

\textbf{Limitations.} We evaluate only probe and full SFT; parameter-efficient methods (LoRA, DoRA) and LP-FT~\cite{kumar2022} may interact differently with channel adaptation and mitigate the negative transfer we observe. With no trained-from-scratch baseline (e.g., EEGNet), the low motor-imagery and emotion accuracies are evidence about transfer, not task ceilings. BENDR uses both the convolutional encoder and the contextualizer~\cite{kostas2021}. FACED remains largely unsolved (best 25.8\%, $\approx 2.3\times$ chance), several cells collapsing near 12.5\%, so it is a failure case rather than fine-grained adapter evidence. Several conditions show high variance (std $>$ 8pp, e.g.\ BENDR SSI/Riemannian SFT on PhysioNet, EEGPT Conv1d on MDD), indicating seed-dependent convergence.

\textbf{A confirmatory experiment.} One controlled study would settle the confound: fix the trunk, corpus, objective, sampling rate, and parameter budget, and vary only the channel front-end: a fixed-montage convolution (rigid) versus a content-conditioned positional encoding and a learned-query bottleneck, pretrained once on a canonical 19-channel montage and once on heterogeneous montages (19--64 channels), then evaluate all five adapters under probe and SFT. The account predicts, falsifiably, that the rigid front-end stays at chance under a probe and needs a learned bridge in \emph{both} regimes; that the flexible front-ends beat a learned adapter natively \emph{only} after heterogeneous-montage pretraining; and that an added external adapter degrades the content-conditioned front-end under SFT but not the bottleneck. A ranking that flips with corpus or scale would refute it. The minimal test is four pretraining runs.

\section{CONCLUSIONS}

We presented the first systematic comparison of channel adaptation methods for EEG foundation models across five models, five datasets, and two training regimes. EEG foundation models do need a channel-adaptation step, but the adapter should preserve the pretrained channel computation rather than overwrite it. A fixed/native route is best in 18 of 25 model--task cells, while regime-specific wins are distributed across all five routes. The strongest model, CBraMod (5M), reaches its best accuracy without any added learned channel adapter, consistent with independent findings~\cite{eegfmworth2026} that compact EEG-specific architectures can rival larger ones. Because learned adapters add montage-dependent task capacity, frozen-probe benchmarks should use a fixed/native route to compare representations; full-SFT results should report the adapter as part of the trained recipe. The remaining choices (rigid vs.\ flexible backbone and probe vs.\ SFT, the latter causing negative transfer in 25.7\% of fine-tuning conditions at the shared learning rate) matter at least as much as the adapter family, though our design still confounds architecture with pretraining data and objectives.

\section*{Acknowledgment}
This work was supported by the NSF (2423943, \#1928224) and the NAIRR program (NAIRR250045), and used the San Diego Supercomputer Center Expanse cluster via ACCESS allocation IBN140002 on the Neuroscience Gateway~\cite{sivagnanam2013nsg, sivagnanam2015nsg}. Claude (Anthropic), an AI assistant, was used under the authors' direction for the reanalysis and Figs.~1--4 (Secs.~IV--V) and the text corrections (abstract, Secs.~I and III--VI); the authors reviewed all changes and take full responsibility for the content.


\end{document}